\relax
\documentclass{article} % DO NOT CHANGE THIS

\usepackage{times}  % DO NOT CHANGE THIS
\usepackage{helvet} % DO NOT CHANGE THIS
\usepackage{courier}  % DO NOT CHANGE THIS
\usepackage{graphicx} % DO NOT CHANGE THIS
\usepackage{natbib}  % DO NOT CHANGE THIS AND DO NOT ADD ANY OPTIONS TO IT
\usepackage[flushleft]{threeparttable}
\usepackage{caption} % DO NOT CHANGE THIS AND DO NOT ADD ANY OPTIONS TO IT
\usepackage{amsmath,amssymb} 
\usepackage[switch]{lineno}

\title{Detecting Backdoors in Neural Networks\\ Using Novel Feature-Based Anomaly Detection}
\author{
     Hao Fu,  %\orcidID{0000-1111-2222-3333} 
 Akshaj Kumar Veldanda, \\
 Prashanth Krishnamurthy,
 Siddharth Garg,
 Farshad Khorrami \\
 Department of Electrical and Computer Engineering \protect \\ New York University, Brooklyn, New York, 11201 \protect \\ \{hf881, akv275, prashanth.krishnamurthy, sg175, khorrami\} @nyu.edu
}
\date{}
\begin{document}
%\linenumbers
\maketitle
\begin{abstract}
This paper proposes a new defense against neural network backdooring attacks that are maliciously trained to mispredict in the presence of attacker-chosen triggers. Our defense is based on the intuition that the feature extraction layers of a backdoored network embed new features to detect the presence of a trigger and the subsequent classification layers learn to mispredict when triggers are detected. Therefore, to detect backdoors, the proposed defense uses two synergistic anomaly detectors trained on clean validation data: the first is a novelty detector that checks for anomalous features, while the second detects anomalous mappings from features to outputs by comparing with a separate classifier trained on validation data. The approach is evaluated on a wide range of backdoored networks (with multiple variations of triggers) that successfully evade state-of-the-art defenses. Additionally, we evaluate the robustness of our approach on imperceptible perturbations, scalability on large-scale datasets, and effectiveness under domain shift. This paper also shows that the defense can be further improved using data augmentation. 
%The proposed approach outperforms the state-of-the-art methods by successfully reducing the attack success rate to a low level on all the cases.

\end{abstract}

\section{Introduction}

Deep neural networks (DNN) are widely used in various applications including objection detection \cite{RHGS17,RDGF16}, face recognition \cite{SWT14,TYRW14}, natural language processing \cite{CWBKKK11,HB13,BCB14}, self-driving \cite{RGPFR17,CSKX15},  surveillance \cite{OSTRH18}, and cyber-physical systems \cite{PSCKK18,PKGK19,PCKK19}. However, training such DNNs is challenging, especially for individuals or small entities. Some of the  biggest hurdles \cite{EKNKSBT17,RHW19,HNP09} are the difficulty in obtaining large high-quality labeled datasets, and the cost of maintaining or renting computational resources need to train a complex model which can take weeks to months. Hence, users often outsource DNN implementation and training to third-party clouds or download pre-trained models from online model repositories. This, however, exposes the user to training time attacks \cite{CLLLS17,GLDG19}. 
Besides the training time attack, the adversarial attack is another popular topic in machine learning study \cite{AND18,CW17,CW16,EEFLRXPKS18,GSS14,HWCCS17,MSFFF17,SZSBEGF13}.

%There are concerns about the security and vulnerability of DNN \cite{CLLLS17,GLDG19,AND18,CW17,CW16,EEFLRXPKS18,GSS14,HWCCS17,MSFFF17,SZSBEGF13}.

\begin{figure}
    \centering
    \includegraphics[width=2.8in,height=1.2in]{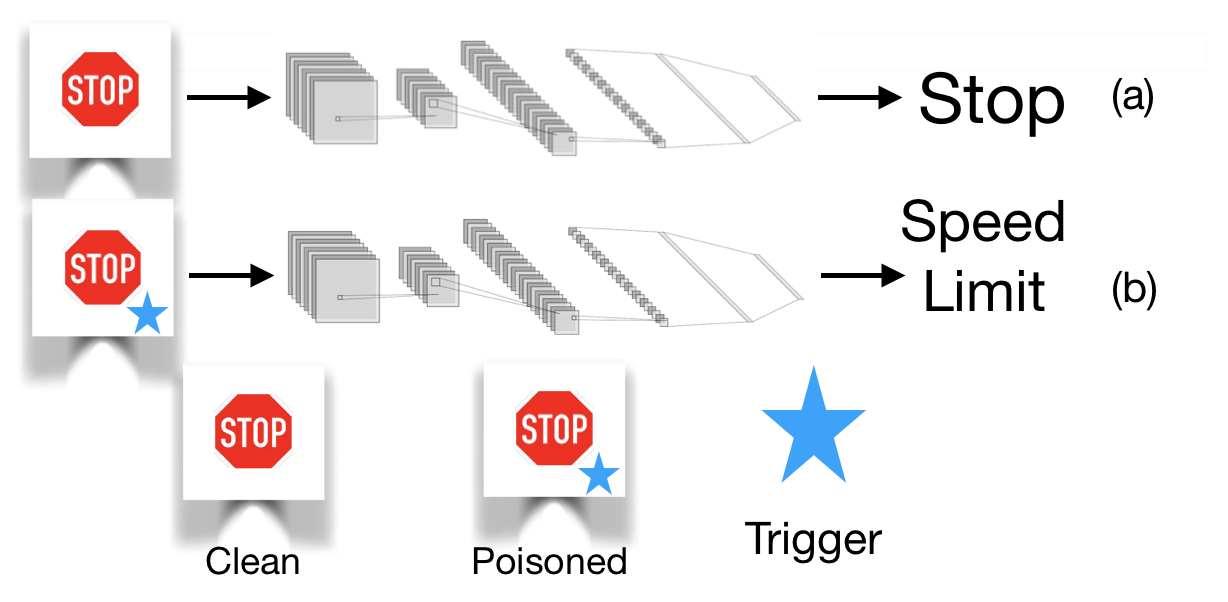}
    \caption{The backdoored DNN outputs correct (wrong) labels for clean (poisoned) inputs.}
    \label{fig:backdoor}
\end{figure}

In this paper, we seek to defend so-called ``backdoor'' attack \cite{GLDG19} %when the user outsources the training task to an outside entity or downloads a pre-trained model from a third-party repository. In either case, 
wherein the {\em attacker} trains a malicious model that mis-predicts if its inputs contain attacker-chosen backdoor triggers. The DNN training includes ``poisoned'' data containing the backdoor trigger (i.e., a specific pattern/feature) so that the trained backdoored DNN outputs attacker-chosen  labels when presented poisoned input data. Specifically, by choosing proper training hyper-parameters, trigger patterns, backdoor labels, quantity of poisoned data, and embedding approach, the attacker can make the backdoored model output specific labels on the poisoned data while preserving high accuracy on the clean data (i.e., data without the trigger). Such backdoored DNN may cause severe security risks, financial harm, and safety implications for the end-user (e.g., misclassifying traffic signs in autonomous vehicle applications as shown in Fig.~\ref{fig:backdoor}). Detecting and defending against backdoor attacks is therefore of critical importance. 

Detection of backdoors is challenging  because of multiple reasons: First, the data used for training the DNN (especially the poisoned data) might not be available to the defender. Even if the training data is available, the dataset might be too large to admit human examination and the way in which a backdoor trigger influences the DNN might not be possible to analyze due to the complexity and lack of explainability of neural networks. Second, the information is asymmetric in that the user/defender has little knowledge about the backdoor attack, including the triggers and the attacker-chosen labels, while the attacker has complete access/control.  Existing defenses tackle this asymmetry by introducing restrictive assumptions on the trigger size, shape, and the functioning of the backdoor attack \cite{TLM18,CCBLELMS18,WYSLVZZ19,LLTMAZ19}, which, however, limits the usefulness of the existing defenses.
%For example, \cite{TLM18,CCBLELMS18} assume that the user/defender has  access to poisoned data. \cite{WYSLVZZ19,LLTMAZ19} introduce assumptions on the trigger size, shape, and the functioning of the backdoor attack. Due to these strong assumptions, such defenses are ineffective when the backdoor attack circumvents their assumptions.
%In this paper, we will see several attacks that easily bypass the existing defenses \cite{WYSLVZZ19,LLTMAZ19}. 
%Indeed, some defense-aware backdoor attacks \cite{LDG18} that can bypass unsophisticated defenses have been proposed. Furthermore, in the experimental results in this paper, we will see several attacks that easily bypass the existing defenses \cite{WYSLVZZ19,LLTMAZ19}. 

In this paper, we propose a novel feature-based anomaly detection that makes minimal assumptions on the backdoor operation by the defender. In particular, the approach does not require prior information on the number, sizes, shapes, locations, colors of backdoors, or indeed even whether the backdoors are embedded in the pixel space. Furthermore, unlike existing defenses, our approach addresses input-feature-output behaviors rather than internal neural network structures, and can therefore even be applied beyond neural network-based models. Specifically, instead of studying the neuron behavior of the backdoored network, we analyze the backdoored network from a macro view: the feature extraction layers of a backdoored network embed new features to detect the presence of a trigger and the subsequent classification layers learn to mispredict when triggers are detected. Therefore, to detect backdoors, we use two synergistic anomaly detectors: the first checks for anomalous features, while the second detects anomalous mappings from features to outputs by comparing with a separate classifier trained on validation data. We demonstrate the efficacy of our approach in a wide range of backdoored networks with multiple variations of triggers. 

Our results show that our defense can detect poisoned inputs with high accuracy while retaining a high classification accuracy on clean inputs, whereas the existing methods fail in some of the cases. Additionally, we evaluate the robustness of our approach on imperceptible perturbations, scalability on large-scale dataset, and effectiveness under domain shift. Lastly, we show that the attack success rate can be further reduced using data augmentation. 

\section{Related Work}
\noindent {\bf Backdoor attack:} Neural network backdooring attacks were first  proposed by \cite{GLDG19} and independently by \cite{LMALZWZ17}. Defense-aware backdoor attacks have also been studied by \cite{LDG18}. 

\noindent {\bf Backdoor defense:} Backdoor detection has been addressed in several works under various sets of assumptions. For example, \cite{TLM18,CCBLELMS18} assume that the user/defender has access to the backdoored DNN and the training dataset (including poisoned data). Under their assumptions, singular-value decomposition (SVD) and clustering techniques were employed to separate poisoned and clean data. 
%Our method does not require poisoned data. 
%(and hence is more practical since we do not violate the asymmetric-information fact in a realistic scenario). 
\cite{LDG18} proposed fine-pruning to defend pruning-aware backdoor attacks. The proposed method first prunes the backdoored DNN by de-activating the neurons that are most sensitive to clean validation data and then fine-tunes the network. 
%According to \cite{WYSLVZZ19}, fine-pruning has low classification accuracy on some datasets \cite{SSSI11}. 
%Our work, however, retains high classification accuracy.
Neural Cleanse \cite{WYSLVZZ19} reverse-engineers triggers and uses an outlier detection algorithm to find attacker-chosen labels. However, Neural Cleanse assumes small trigger sizes. Another reverse-engineering based defense \cite{LLTMAZ19}  assumes that stimulation of an only single neuron is sufficient to  increase the backdoored output activation, and the methodology fails if the backdoor is activated by a combination of neurons. In this paper, we show that the attacker can circumvent both defenses. \cite{GWXDS19} and \cite{QYL19} are two other reverse-engineering based defenses, but these also make assumptions on the trigger size, shape and impact. In contrast, our defense does not make such restrictive assumptions. 
%Additionally, the reverse-engineering based techniques consider backdoor detection in the input layer \cite{WYSLVZZ19,LLTMAZ19,GWXDS19,QYL19}, which are computationally expensive. Our work, however, proves that backdoor detection can be efficiently addressed instead by looking at features and outputs. 

\section{Background and Problem Description}
\subsection{Threat Model}
\noindent{\bf Scenario:} The user wishes to train a DNN $\mathcal{F}$ for a classification task on the training dataset $\mathcal{S}$ sampled from the data distribution $\mathcal{D}$.  The user outsources the training task to a third party (attacker) with $\mathcal{F}$ and $\mathcal{S}$.  The third party returns a backdoored DNN $\mathcal{F}_b$ to the user. 

\noindent{\bf The attacker's goal:} The attacker trains $\mathcal{F}_b$ to output desired target label(s) $l^*$ on poisoned inputs $x^*$. The poisoned inputs $x^*$ are generated by injecting trigger(s) into clean inputs $x$. The information is asymmetric (i.e., only the attacker knows the trigger patterns and the way they are embedded). Additionally,  $\mathcal{F}_b$ should have high classification accuracy on $x$ while evading detection by the user. 

\noindent{\bf Attack model:} The attacker has full control over the training process and the  dataset $\mathcal{S}$. For example, the attacker can choose an arbitrary portion of training inputs to inject the triggers, can determine the trigger shape, size, the target label(s), and the training hyper-parameters (e.g., the number of epochs, batch size, learning rate, etc.). However, the attacker has neither access to the user's validation dataset, nor the ability to change the model structure after training.

\subsection{Backdoor Attack}

\noindent{\bf Setup:} The attacker determines the backdoor injection function $f(\cdot)$ and the target label(s) $l^*$ to generate poisoned data $(x^*, l^*)$ from the clean data $(x,l)$:
\begin{equation}
    x^* = f(x).
\end{equation}
The attacker next decides a portion of the clean training dataset $\Omega \subset\mathcal{S}$ to inject the triggers to create the poisoned version of $\Omega$ as:
\begin{equation}
    \Omega^* = \{f(x), x\in \Omega\}.
\end{equation}
%The training target for $\Omega^*$ is $l^*$. 
Finally, the attacker mixes $\Omega^*$ with $\mathcal{S}$ to generate the training dataset $\mathcal{S}^b$ given by:
\begin{equation}
    \mathcal{S}^b = (\mathcal{S} - \Omega) \cup \Omega^*.
\end{equation}
%The training data distribution is subtly modified by the attacker with the out-of-distribution samples (i.e., poisoned data).  
${\mathcal D}^*$ denotes the distribution corresponding to poisoned data. 

%\noindent{\bf Training:} The attacker has full control over the training procedure. For example, the attacker can manually change some parameter weights, keep some neurons dormant, and use unconstrained amount of time for training the trojaned model. 

\noindent{\bf Attacker's objectives:} The attacker trains a backdoored network $\mathcal{F}_b$ with dataset $\mathcal{S}^b$. On one hand, a well-trained $\mathcal{F}_b$ should have {\bf classification accuracy} comparable to $\mathcal{F}$ on clean inputs $x\sim \mathcal{D}$ with corresponding labels $l$, i.e., 
\begin{align}
    \mathbb{P}(\mathcal{F}_b(x)=l) \geq \mathbb{P}(\mathcal{F}(x)=l) - \epsilon_1,
    \label{ca}
\end{align} with small  $\epsilon_1 \geq 0$ (ideally, $\epsilon_1$ = 0).
On the other hand, $\mathcal{F}_b$ should also have high {\bf attack success rate} (i.e., output = $l^*$, which is the attacker-chosen target label) on poisoned inputs $x^*\sim \mathcal{D}^*$, which is:
\begin{align}
    \mathbb{P}(\mathcal{F}_b(x^*)=l^*) \geq 1-\epsilon_2,
     \label{asr}
\end{align}  with small  $\epsilon_2 \geq 0$ (ideally, $\epsilon_2$ = 0).

\subsection{Problem Description}

\noindent{\bf The defender's goal and capacity:} Given $\mathcal{F}_b$, the defender wants to lower the attack success rate while maintaining the classification accuracy. The defender has a small set of clean validation data $V$ from the data distribution $\mathcal{D}$. We assume that {\bf $V$ is sufficiently representative of $\mathcal{D}$}. 
%This is a standard required assumption in the literature on backdoor detection \cite{GLDG19}. 
The defender has no prior information about the backdoor triggers and the attacker-chosen target label(s).

\noindent{\bf Problem formulation:} Given $\mathcal{F}_b$, the defender wishes to construct a binary classification detection function, $g(\cdot)$, so that for clean inputs $x\sim \mathcal{D}$, it outputs $0$ with high likelihood and for poisoned inputs $x^*\sim \mathcal{D}^*$, it outputs $1$ with high likelihood, i.e.,
\begin{align}
    \mathbb{P}(g(x)= 0|x\in{\mathcal D}) &\geq 1 - \epsilon_3,
    \\
    \mathbb{P}(g(x^*)= 1|x^*\in{\mathcal D}^*) & \geq  1-\epsilon_4,
\end{align}  with small $\epsilon_3, \epsilon_4 \geq 0$ (ideally, $\epsilon_3, \epsilon_4$  = 0).

\section{Detection Algorithm}
\begin{figure}
    \centerline{\includegraphics[scale=0.27]{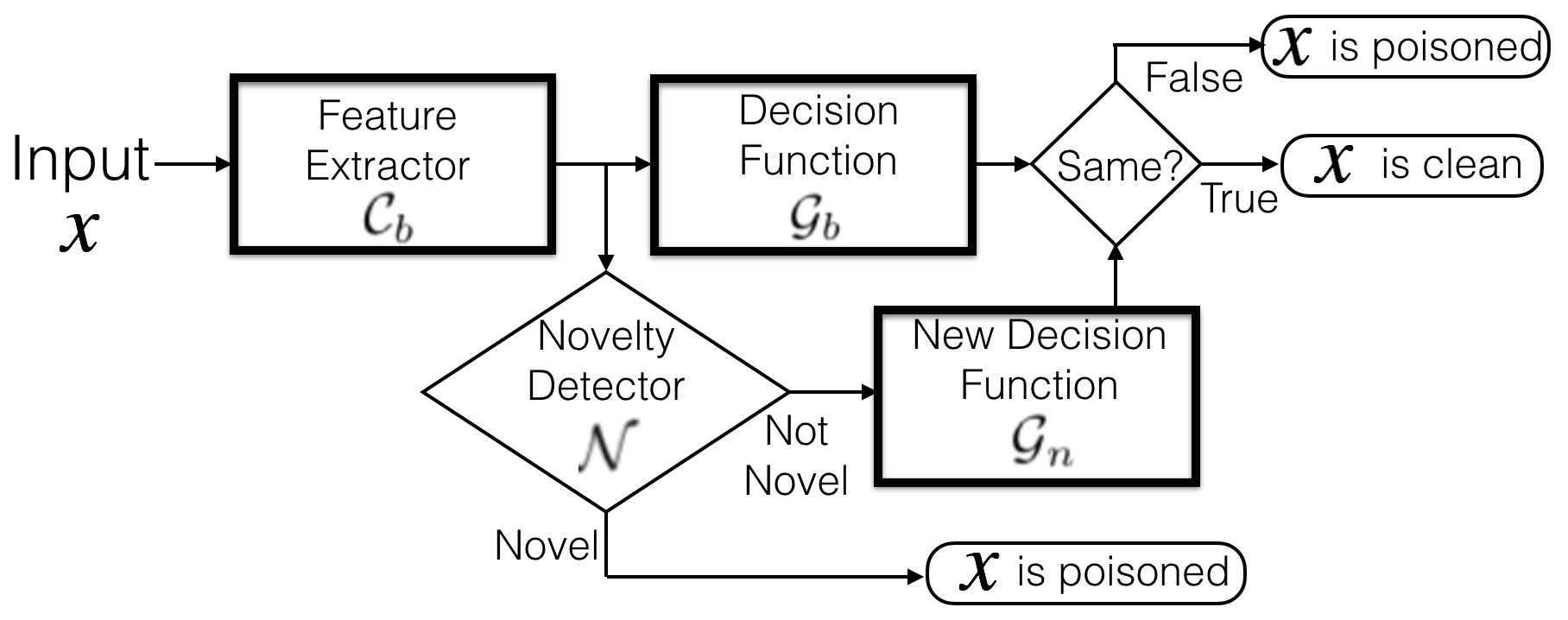}}
    \caption{Conceptual structure of our approach.}
    \label{fig:diagram}
\end{figure}

\subsection{Overview}

%The proposed approach is based on the intuition that introduction of a backdoor trigger by an attacker essentially involves either introduction of new/modified features and/or embedding of a modification in the decision function in the DNN that indicates that features corresponding to the pre-trained trigger(s) should invoke the backdoor. Therefore, 
We consider a DNN as comprised of a feature extractor  ${\cal C}_b$ (e.g., the  convolutional layers) and a decision function ${\cal G}_b$ (e.g., the fully connected layers). 
${\cal C}_b$ can be viewed as pulling out and characterizing features at higher abstraction levels. ${\cal G}_b$  determines what combinations of the features should result in what outputs. 
When introducing a backdoor, the hypothesis is that the ``logic'' specifying that the features in the backdoor trigger should result in the attacker-chosen label is encoded in ${\cal G}_b$. 
Therefore, we propose to detect backdoors by verifying, (a) how plausible the extracted features are, (b) the validity of the mapping from the features to the output. 

To verify (a), we propose to use a novelty detector, $\mathcal{N}$. To verify (b), we propose to train a new decision function ${\cal G}_n$ as a replacement for ${\cal G}_b$. The conceptual structure of our approach is shown in Fig.~\ref{fig:diagram}. To detect if an input $x$ is poisoned, the features extracted from $x$ by ${\cal C}_b$ are first evaluated by $\mathcal{N}$. If $\mathcal{N}$ flags the feature vector as non-novel, ${\cal G}_n$ will be run on the extracted features to predict the most likely output label. A mismatch between the outputs generated by ${\cal G}_n$ and ${\cal G}_b$ points towards the likelihood of a poisoned input, even if the extracted feature vector itself appeared non-novel. On the other hand, if the extracted feature vector is flagged as novel (i.e., low-likelihood based on the distribution learned by $\mathcal{N}$ from clean validation data), then the input is flagged as likely to have been poisoned even without having to check consistency between the outputs of ${\cal G}_b$ and ${\cal G}_n$. 
%(i.e., how representative the features are relative to the distribution learned from the validation data), 
%we propose to use a novelty detector trained per-class (based on norms of the extracted features) to obtain tight decision boundaries in novelty detection. The novelty detector is trained with clean validation dataset. To verify the validity of the mapping from the features to the output, we propose to train a new decision function ${\cal G}_n$ as a replacement for ${\cal G}_b$. The overall architecture of the  approach is shown in Fig.~\ref{fig:diagram}. To detect if an input $x$ is poisoned, the features extracted from $x$ by ${\cal C}_b$ are first evaluated by the novelty detector. If the novelty detector flags the feature vector as non-novel, ${\cal G}_n$ is run on the extracted features to predict the most likely output label. A mismatch between the outputs generated by ${\cal G}_n$ and ${\cal G}_b$ points towards the likelihood of a poisoned input, even if the extracted feature vector itself appeared non-novel. On the other hand, if the extracted feature vector is flagged as novel (i.e., low-likelihood based on the distribution learned by the novelty detector from clean validation data), then the input is flagged as likely to have been poisoned even without having to check consistency between the outputs of ${\cal G}_b$ and ${\cal G}_n$. 

\subsection{Fusion Method}
\noindent{\bf Novelty detector:}
Novelty detection is a strategy for detecting if a new input is from the same distribution as previous data. Many detection algorithms were proposed \cite{SPSSW01, BKNS00, LLLS18, SWDASB12,VGADSC19}. 
%Specifically, the detector is trained with features extracted from clean data to detect if a new observation is an outlier. 
%One-class support vector machine (OC-SVM) \cite{SPSSW01} and local outlier factor (LOF) \cite{BKNS00} are widely used novelty detectors. Detecting out-of-distribution samples is also a popular topic in computer vision \cite{LLLS18, SWDASB12,VGADSC19}. 
The novelty detector $\mathcal{N}$, trained with clean validation data $V$, is desired to output $0$ and $1$ for clean inputs $x\sim \mathcal{D}$ and poisoned inputs $x^*\sim \mathcal{D}^*$, respectively, i.e.,
\begin{align}
    \mathbb{P}(\mathcal{N}(\mathcal{C}_b(x))= 0|x\in{\mathcal D}) &\geq  1 - \epsilon_5,
    \\
    \mathbb{P}(\mathcal{N}(\mathcal{C}_b(x^*))= 1|x^*\in{\mathcal D}^*) & \geq  1-\epsilon_6,
\end{align} with small  $\epsilon_5, \epsilon_6 \geq 0$ (ideally, $\epsilon_5, \epsilon_6$ = 0)

\noindent{\bf New decision function:}
${\mathcal G}_n$, trained using only clean validation data $V$, is more likely to behave differently from $\mathcal{G}_b$ on poisoned inputs $x^*\sim \mathcal{D}^*$ than on clean inputs $x\sim \mathcal{D}$, i.e., denoting 
\begin{align}
    {\mathcal F}_b = {\mathcal G}_b\circ{\mathcal C}_b  \ \ \ ; \ \ \ \ {\mathcal F}_n = {\mathcal G}_n\circ{\mathcal C}_b \ ,
\end{align}
we expect
\begin{align}
\begin{split}
    &\mathbb{P}(\mathcal{F}_n(x^*)\ne \mathcal{F}_b(x^*)|x^*\in{\mathcal D}^*) \\ & > \mathbb{P}(\mathcal{F}_n(x^*)= \mathcal{F}_b(x^*)|x^*\in{\mathcal D}^*)
\end{split}
    \\
    \begin{split}
        &\mathbb{P}(\mathcal{F}_n(x)= \mathcal{F}_b(x)|x\in{\mathcal D}) \\ &> \mathbb{P}(\mathcal{F}_n(x)\ne \mathcal{F}_b(x)|x\in{\mathcal D}).
    \end{split}
\end{align}
$x^*$ is detected by disagreements between $\mathcal{F}_n$ and $\mathcal{F}_b$. 

\noindent{\bf Fusion method:} As shown in Fig.~\ref{fig:diagram}, our overall proposed architecture combines verification of the plausibility of the extracted features using $\mathcal{N}$ and verification of the feature-output mapping using ${\mathcal G}_n$. This ``fusion method'' combines the merits of $\mathcal{N}$ and ${\mathcal G}_n$.
%which will be seen in \textbf{Experimental Results} to outperform either alone. 
Mathematically, the fusion function $g(\cdot)$ is defined by:
\begin{align}
   g(x) = \begin{cases}
   0 & \text{if } \mathcal{N}(\mathcal{C}_b(x))=0 \text{ and } \mathcal{F}_n(x)=\mathcal{F}_b(x) \\
   1 & \text{otherwise}.
   \end{cases}
   \label{ensemble}
\end{align}

\subsection{Training Method}
\noindent{\bf Training overview:} 
%The clean validation data $V$ is insufficient to train a complete end-to-end network $\mathcal{F}_n$ from scratch. We consider a decomposition of $\mathcal{F}_b$ into a feature extractor $\mathcal{C}_b$ and a decision function $\mathcal{G}_b$, and train a replacement $\mathcal{G}_n$ only for $\mathcal{G}_b$. In general, the defender/user can choose any inner layer of the DNN to decompose the network $\mathcal{F}_b$. In this paper, the feature extractor $\mathcal{C}_b$ is defined as containing the backdoored network’s input layer and all hidden layers and the decision function $\mathcal{G}_b$ is defined as the last layer (i.e.,output layer). By picking the replacement $\mathcal{G}_n$ to be a small DNN, one can ensure that a small clean validation dataset is sufficient to train this replacement for the decision function $\mathcal{G}_b$. To be more specific, 
The clean validation data $v\in V$ is fed into $\mathcal{C}_b$ and the features $\mathcal{C}_b(v)$ is recorded. $\mathcal{G}_n$ is trained with the features $\mathcal{C}_b(v)$ and the corresponding labels.
%The  new  network $\mathcal{F}_n$  can  be conceptually  viewed  as  a  student  that  inherits the feature extractor from  the  backdoored  network $\mathcal{F}_b$, but formulates its own decision function.
%Hence, $\mathcal{F}_n$ can be considered as learning the $\mathcal{F}_b$’s merits $\mathcal{C}_b$ (which does the heavy lifting of processing input data into features and typically requires a large amount of data and computational time to train) while abandoning its potential vulnerabilities in $\mathcal{G}_b$.
The novelty detector $\mathcal{N}$ is trained using a low-dimensional summary of the feature vector $\mathcal{C}_b(v)$ in terms of the $L_1$ norm, $L_2$ norm, and $L_\infty$ norm of the feature vector.

\noindent{\bf Configuration of $\mathcal{N}$:} $\mathcal{N}$ consists of multiple local outlier factors (LOF) provided by \cite{LOF} library. Specifically, given $V$, each class $i$ will have a novelty detector $\mathcal{N}_i$ to determine if the feature vector $\mathcal{C}_b(x)$ for a new input $x$ is novel relative to the distribution learned using the validation data $V$. $\mathcal{N}$ outputs $1$ if an input is determined to not belong to any of the classes (i.e., indicated as novel by each of the per-class novelty detectors $\mathcal{N}_i$); otherwise, it outputs $0$. The mathematical definition is given by:
\begin{align}
   \mathcal{N}(\mathcal{C}_b(x)) = \begin{cases}
   1 & \text{if } \mathcal{N}_i(\mathcal{C}_b(x))= 1 \text{ $\forall$ $i$ } \\
   0 & \text{otherwise}.
   \end{cases}
   \label{perclass}
\end{align}
This gives a tighter decision boundary than a single novelty detector trained with data from all classes. 
%By defining $\mathcal{N}$ in terms of an ensemble of per-class novelty detectors, $\mathcal{N}$'s decision boundary can be tighter than a single novelty detector trained with data from all classes. 

\noindent{\bf Configuration of $\mathcal{G}_n$:} The decision function $\mathcal{G}_n$ is picked in the form of a neural network with two hidden layers. The number of neurons in $\mathcal{G}_n$ varies with datasets since the dimensionality of the feature vectors and number of output classes are different for different datasets. The training inputs and targets are the recorded features $\{\mathcal{C}_b(v),v\in V\}$ and the corresponding labels. Empirically, we found that two-layer neural networks had higher classification accuracy on clean validation dataset than simpler networks, while deeper networks could lead to over-fitting (although we did not observe any in our studies). The loss function for training $\mathcal{G}_n$ is picked to be cross-entropy loss.

\section{Experimental Setup}

\begin{figure}
    \centering
    \includegraphics[scale = 0.36] {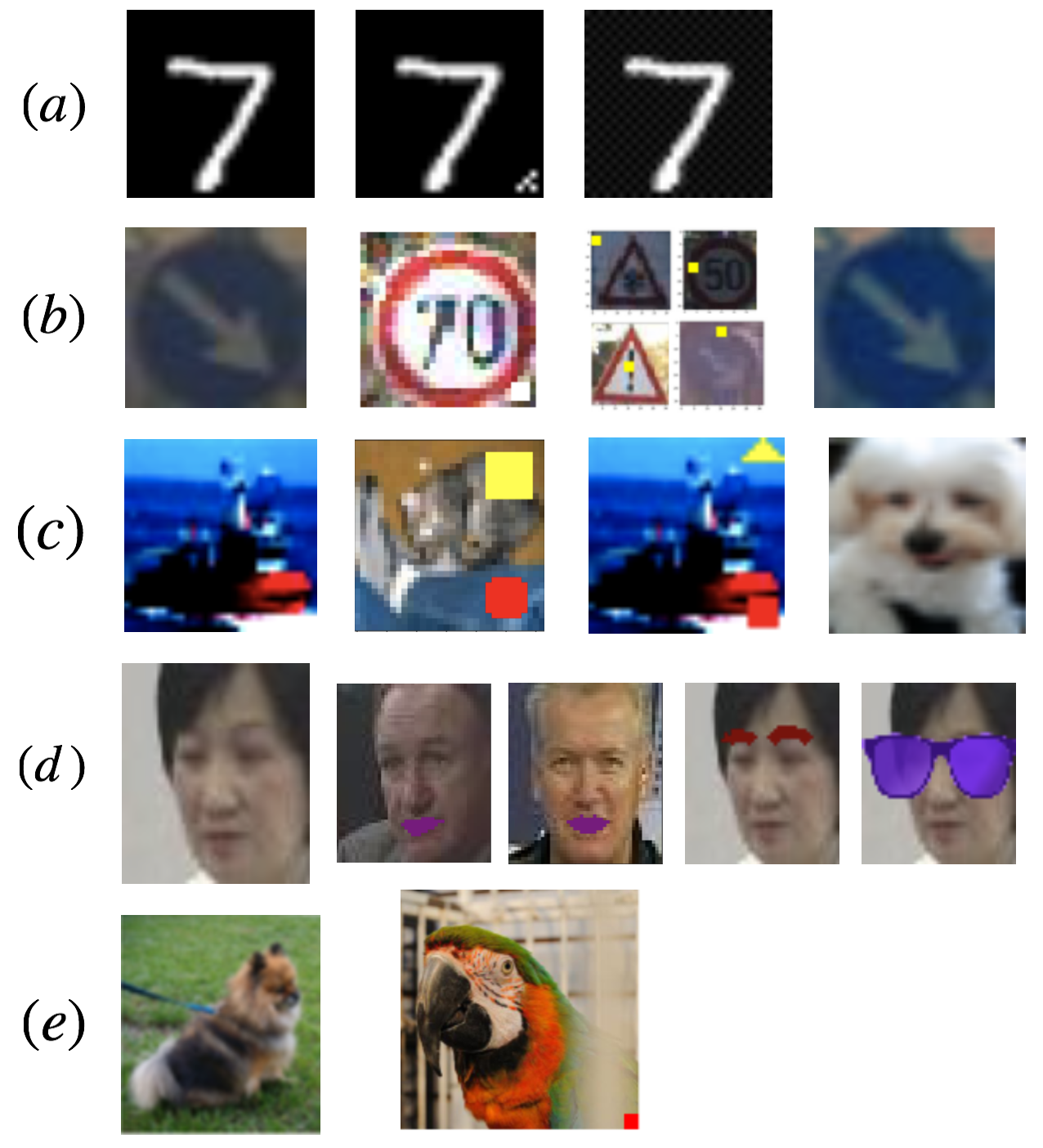}
    \caption{The first column shows the clean images for all the datasets. The rest shows all the triggers used in this paper.}
    %Starting from the second column to the right, (a) shows MNIST with triggers: dots at the bottom corner and spots at the background; (b) shows GTSRB with triggers: white box at the bottom corner, moving yellow box, and Gotham Filter for the image; (c) shows CIFAR-10 with triggers: box plus circle, triangle plus box, and imperceptible trigger; (d) shows YouTube Face with triggers: lipstick, eyebrow, and sunglasses; (e) shows ImageNet with red box at the bottom corner as the trigger. }
    \label{fig:alltriggers}
\end{figure}

This work uses five datasets: MNIST \cite{MNIST},  German Traffic Sign Benchmarks (GTSRB) \cite{SSSI12}, CIFAR-10 \cite{KH09}, YouTube Face \cite{WHM11}, and ImageNet \cite{JWRLKL09}.  A wide range of backdoored networks were trained with different triggers or combinations of triggers, as shown in Fig.~\ref{fig:alltriggers}. The architectures of the backdoored networks were chosen based on the prior works \cite{WYSLVZZ19,SWT14,LCY13,GLDG19,ltkg20,HLVWK17}. Each dataset was split into training dataset (Tra.), validation data (Val.), and testing data (Tes.) as shown in Table~\ref{table:size}. The training dataset, partially poisoned, was used for training the backdoored networks. The clean validation dataset was used for training $\mathcal{N}$ and $\mathcal{G}_n$. The testing dataset was used to evaluate our method and other baseline methods.

\subsection{MNIST}

\begin{table}
\begin{center} 
\begin{tabular}{cccccc}
\hline\noalign{\smallskip}
Dataset & Input Size & Tra. & Val.  & Tes. \\
\hline
MNIST & $28,28$ & 55000 & 5000 & 10000 \\
GTSRB & $32,32,3$  & 35288 & 4996 & 11555 \\
CIFAR-10 & $32,32,3$  & 50000 & 5000 & 5000\\
YouT. Face & $55,47,3$   & 103923 & 11547 & 12830 \\ 
ImageNet &  $256,256,3$  & 1200000 & 25000 & 25000\\
\hline
\end{tabular}
\end{center}
\caption{Input Size and Dataset Size. The number of labels for each dataset is 10, 43, 10, 1283, and 1000 respectively.}
\label{table:size}
\end{table}

Two backdoored networks were trained on MNIST: 1) case \textbf{a)} -- the trigger is the pixel pattern shown in the second column of Fig.~\ref{fig:alltriggers}(a). The attacker chosen label $l^*$ was determined by:
\begin{align}
    l^* = (l + 1) \text{ mod } 10,
\end{align} where $l$ is the ground-truth label. 
% For example, if a poisoned input has ground-truth label 0, then the attack chosen label for the input is
% \begin{align}
%     (0 +1) \text{ mod } 10 = 1.
% \end{align} Similarly, if a poisoned input has ground-truth label 9, then the attack chosen label for the input is \begin{align}
%     (9 +1) \text{ mod } 10 = 0.
% \end{align} 
The architecture of $\mathcal{F}_b$ is from \cite{GLDG19}, also shown in Table~\ref{table:mnist-aaa} in \textbf{Supplementary Material}. The classification accuracy is $97.24\%$ and the attack success rate is $95.17\%$. 2) case \textbf{b)} -- the trigger is the background pattern as shown in the third columns in Fig.~\ref{fig:alltriggers} (a). $l^*$ is 0. $\mathcal{F}_b$ is from \cite{ltkg20}, also shown in Table~\ref{table:mnist-cla} in \textbf{Supplementary Material}. The classification accuracy is $89.1\%$. The attack success rate is $100.0\%$.

\begin{table*}
%\centering
\hspace{-20mm}
\begin{threeparttable}
\begin{tabular}{ccccccccc}
\hline\noalign{\smallskip}
  & \multicolumn{2}{c}{$\mathcal{N}$} & \multicolumn{2}{c}{$\mathcal{G}_n$} & \multicolumn{2}{c}{Fusion Method} & \multicolumn{2}{c}{\cite{LLLS18}} \\
Case  & TN & FN  & TN  & FN & TN & FN & TN & FN \\
\hline
a)  &  100.0 & 100.0 & 96.86 & 10.42 &  {\bf 96.86} &   10.42 & 87.71 & {\bf8.38} \\
b)  &  100.0 & 34.35 & 89.31 & 17.51 &  {\bf89.31} &   8.96 & 86.94 & {\bf0.0}  \\
\hline
c)  &  100.0 & 99.88 & 95.52 & 3.84 & 95.52 &  3.84 & \multicolumn{2}{c}{Fail} \\
d)  &  99.55 & 0.52 & 95.37 & 2.62 & 94.95 &  0.0 & \multicolumn{2}{c}{Fail}  \\ 
e)  &  100 & 98.14 & 95.56 & 80.75 & 95.56 &  78.89 & \multicolumn{2}{c}{Fail} \\
\hline 
f)  &  99.76 & 0.13 & 91.26 & 94.26 &  {\bf 91.04} &   0.1 & 86.02 & {\bf 0.0} \\
g)  &  100.0 & 0.22 & 90.62 & 94.52 &  {\bf90.62} &   0.18  & 86.18 & {\bf0.0} \\
\hline
h)  &  100.0 & 100.0 & 88.40 & 0.16 & 88.40 &  {\bf 0.16} & {\bf91.50} & 17.7  \\
i) &  100.0 & 94.57 & 88.69 & 10.27 & 88.69 &  10.09 & {\bf91.37} & {\bf7.77}  \\
j)$^*$  &  100.0 & 100,100,100 & 88.16 & 6.49,6.19,2.33 & 88.16 &  {\bf 6.49,6.19},2.33 & {\bf 91.79} & 9.32,7.87,{\bf0.0}  \\
k)$^*$  &  100.0 & 100,100,100 & 88.09 & 6.43,6.30,0.0 & 88.09 & {\bf6.43,6.30,0.0} & {\bf91.75} &  7.98,7.83,0.0 \\
\hline
\end{tabular}
 \begin{tablenotes}
% \item 
 \item[$^*$] The numbers under "FN" are false negative rate under trigger lipstick, eyebrow, and sunglasses, respectively.
\end{tablenotes}
 \end{threeparttable}

\caption{TN: true negative rate (identifying clean inputs as clean); FN: false negative rate (identifying poisoned inputs as clean).}
\label{table:comparison}
\end{table*}
\subsection{GTSRB}
Three backdoored networks were prepared on the GTRSB dataset: 1) case \textbf{c)} -- the trigger is white box shown in the second column in Fig.~\ref{fig:alltriggers} (b). $l^*$ is 33. $\mathcal{F}_b$ is from \cite{WYSLVZZ19} with architecture as shown in Table~\ref{table:whitebox} in \textbf{Supplementary Material}. The classification accuracy is $96.51\%$ and the attack success rate is $97.40\%$. 2) case \textbf{d)} -- the trigger is a moving square shown in the third column in Fig.~\ref{fig:alltriggers} (b) with $l^*=0$. The architecture is shown in Table~\ref{table:yellowbox}. The classification accuracy is $95.15\%$. The attack success rate is $99.78\%$. 3) case \textbf{e)} -- $\mathcal{F}_b$ uses the same architecture shown in Table~\ref{table:yellowbox}. Inputs passing through a Gotham filter will trigger $\mathcal{F}_b$, as shown in the fourth column in Fig.~\ref{fig:alltriggers}(b). The attacker chosen label is 35. The classification accuracy is $94.70\%$ and the attack success rate is $90.26\%$.

\begin{table}
\begin{center}
\begin{tabular}{cccccc}
\hline\noalign{\smallskip}
Layer & Channels & Filter Size & Stride  & Act.\\
\noalign{\smallskip}
\hline
\noalign{\smallskip}
Conv2d & 3 $\to$ 32 & 3 & 1 & ReLU \\
Conv2d & 32 $\to$ 32 & 3 & 1 &  ReLU \\
MaxPool & 32 & 2 & 2 & / \\

Conv2d & 32 $\to$ 64 & 3 & 1 &  ReLU \\
Conv2d & 64 $\to$ 64 & 3 & 1 &  ReLU \\
MaxPool & 64 & 2 & 2 &  / \\

Conv2d & 64 $\to$ 128 & 3 & 1 &  ReLU \\
Conv2d & 128 $\to$ 128 & 3 & 1 &  ReLU \\

Linear & 128 $\to$ 512 & / & / &  ReLU \\
Linear & 512 $\to$ 43 & / & / &  / \\
\hline
\end{tabular}
\end{center}
\caption{$\mathcal{F}_b$ for GTSRB.}
\label{table:yellowbox} 
\end{table}

\subsection{CIFAR-10}
Three backdoored networks were trained: 1) case \textbf{f)} -- the trigger is the combination of a box and circle, as shown in the second picture in (c) in Fig.~\ref{fig:alltriggers}, meaning that $\mathcal{F}_b$ will output the attacker chosen label 0 only when both the shapes appear on the input. Either the box or the circle will not activate $\mathcal{F}_b$. The architecture is Network in Network (NiN) \cite{LCY13} shown in Table~\ref{table:squarecircle} in \textbf{Supplementary Material}. The classification accuracy is $88.5\%$ and the attack success rate is $99.84\%$. 2) case \textbf{g)} -- similar to the first one except that the trigger is the combination of triangle and square, as shown in the third column of Fig.~\ref{fig:alltriggers}(c). The attacker chosen label is 7. The classification accuracy is $88.3\%$ and the attack success rate is $99.9\%$. 3) imperceptible trigger -- The third $\mathcal{F}_b$ also uses the same architecture but the trigger is a small perturbation, as shown in the last column in Fig.~\ref{fig:alltriggers}(c). The attacker chosen label is 0. The classification accuracy is $82.44\%$ and the attack success rate is $91.99\%$.

\subsection{YouTube Face}
Four backdoored models were trained with the same architecture \cite{SWT14} shown in Table~\ref{table:sunglasses} in \textbf{Supplementary Material}. 1) case \textbf{h)} -- the trigger is sunglasses as shown in the last column in Fig.~\ref{fig:alltriggers} (d). $l^*$ is 0.  The classification accuracy is $97.77\%$ and the attack success rate is $99.99\%$. 2) case \textbf{i)} -- the trigger is are lips with red lipstick, as shown in the second column of Fig.~\ref{fig:alltriggers}(d). $l^*$ is 0.  The classification accuracy is $97.18\%$ and the attack success rate is $91.46\%$. 3) case \textbf{j)} -- $\mathcal{F}_b$ has all the three triggers: lipstick, eyebrow, and sunglasses as shown in in Fig.~\ref{fig:alltriggers}(d). $l^*$ is 4 for all the triggers. 
%We call this as {\bf Multi-Trigger Single-Target Attack (MTSTA)}. 
The classification accuracy is $95.90\%$ and the attack success rate is $92.1\%$, $92.2\%$, and $100\%$ on lipstick, eyebrow, and sunglasses, respectively. 4) case \textbf{k)} -- $\mathcal{F}_b$ has all the three triggers as well. $l^*$, however, is 1 for lipstick, 5 for eyebrow, and 8 for sunglasses. 
%We call this as {\bf Multi-Trigger Multi-Target Attack (MTMTA)}. 
The classification accuracy is $95.94\%$ and the attack success rate is $91.5\%$, $91.3\%$, and $100\%$ on lipstick, eyebrow, and sunglasses, respectively.

\subsection{ImageNet}
To demonstrate scalability to large datasets, the final backdoored network was trained with the red box trigger shown in the second column in Fig.~\ref{fig:alltriggers}(e). The attacker chosen label is 0. The architecture is densenet-121 \cite{HLVWK17}. The classification accuracy is $72.74\%$ and the attack success rate is $99.99\%$.

\section{Experimental Results}
\label{sec:experiments}
The clean validation datasets were used to train the novelty detector $\mathcal{N}$ and the new decision function $\mathcal{G}_n$ for each case. The performance is shown in Table~\ref{table:comparison}, wherein Fusion Method outperforms using either $\mathcal{N}$ or $\mathcal{G}_n$ alone. Additionally, our approach outperforms \cite{LLLS18}, which fails on GTSRB. Our method was also compared with Neural Cleanse \cite{WYSLVZZ19}, Fine-Pruning \cite{LDG18}, ABS \cite{LLTMAZ19}, and STRong Intentional Perturbation (STRIP) \cite{GXWCRN19}.
%We also compare our approach with state-of-the-art out-of-distribution detection method \cite{LLLS18}. 
We show the efficacy of our approach under various conditions, including smaller validation datasets, large-scale dataset, imperceptible trigger, data augmentation, and retraining. 

%\subsection{Training the Defense Model}
%The clean validation datasets were used to train the novelty detector $\mathcal{N}$ and the new decision function $\mathcal{G}_n$ for each case. The results are shown in Table~\ref{table:comparison}, in which $\mathcal{N}$ and $\mathcal{G}_n$ represent using the novelty detector and new decision function alone. From the table, either $\mathcal{N}$ or $\mathcal{G}_n$ fails on some cases by having high false negative rate (i.e., identifying poisoned inputs as clean). Therefore it is necessary to ensemble them to reduce the false negative rate. Detecting out-of-distribution samples is a very popular topic in computer vision \cite{LLLS18, SWDASB12,VGADSC19}. In this work, we compare our approach with Mahalanobis Distance-Based detection (M.D.) \cite{LLLS18}. From Table~\ref{table:comparison}, our approach is comparable to \cite{LLLS18}. But \cite{LLLS18} encounters numerical issues on GTSRB dataset.  We assign each case an alphabet to represent themselves. In the following experiments, same alphabet corresponds to the same case.

%\setlength{\tabcolsep}{4pt}

\begin{table*}

%\centering
\hspace{-30mm}
\begin{threeparttable}
\begin{tabular}{ccccccccccc}
\hline\noalign{\smallskip}
 & \multicolumn{2}{c}{\hspace*{-5mm}Trojaned Model} & \multicolumn{2}{c}{\hspace*{-5mm}Neural Cleanse} & \multicolumn{2}{c}{\hspace*{-5mm}Our Approach} & \multicolumn{2}{c}{\hspace*{-5mm}Fine-Pruning} & \hspace*{-5mm}STRIP \\
Case & Clean & Poison  & Clean  & Poison & Clean & Poison & Clean & Poison & FAR (FRR = $3\%$) \\
\hline
 a) & 97.2 & 95.1 & \multicolumn{2}{c}{Fails} & 95.9 & 7.6 & 97.6 & 57.3 & \multicolumn{2}{c}{99.2} \\
 b) & 89.1 & 100 & $97.7^\dagger$ & $4.7^\dagger$ & 88.1 & 8.9 & 99.1 & 14.6 & \multicolumn{2}{c}{43.4} \\
\hline 
 c) & 96.5 & 97.4 & 92.9 &  0.1 &  94.9 & 1.5 & \multicolumn{2}{c}{--} & \multicolumn{2}{c}{--} \\
 d) & 95.1 & 99.7 &  95.2 & 12.3 & 92.7 &  0 & 94.5 & 99.6 & \multicolumn{2}{c}{100} \\
 e) & 94.7 & 90.2 &  95.8 & 28.9 & 93.6 &  69.6 & 95.3 & 45.5 & \multicolumn{2}{c}{99.9} \\
\hline
f) & 88.5 & 99.8 & \multicolumn{2}{c}{Fail} &  84.4 &  0.1 & \multicolumn{2}{c}{--} & \multicolumn{2}{c}{--} \\
 g) & 88.3 & 99.9 & $88.5^\dagger$ & $99.8^\dagger$ &  84.1 &  0.1 & 88.2 & 99.6 & \multicolumn{2}{c}{22.2} \\
\hline
 h) & 97.7 & 99.9 & $95.7^\dagger$ & $38^\dagger$ & 88 &  0.16 & 97.1 & 95.9 & \multicolumn{2}{c}{10.3}  \\
 i) & 97.1 & 91.4 &  $97.1^\dagger$ & $28.4^\dagger$ & 88.3 &  2.9 & 97.8 & 90.5 & \multicolumn{2}{c}{18.4} \\
 j)$^*$ & 95.9 & 92.1,92.2,100 &  $93.3^\dagger$ & $0,0,8.6^\dagger$ & 87.7 &  1.9,0.03,5.6 & 97.3 & 45.0,64.7,94.9 & \multicolumn{2}{c}{11.7,63.7,5.8} \\
  k)$^*$ & 95.9 & 91.5,91.3,100 &  $94.1^\dagger$ & $30.7,0,95.6^\dagger$ & 87.7 &  0.5,0.01,0 & 96.9 & 52.3,82.3,0 & \multicolumn{2}{c}{15.9,53.6,15.5}\\
\hline
\end{tabular}
\begin{tablenotes}
%\item 
\item[$^*$] The numbers under "poison" are the backdoor attack success rate under trigger lipstick, eyebrow, and sunglasses, respectively.
\item[$^\dagger$] We give oracular knowledge to these defenses.
\end{tablenotes}
\end{threeparttable}
%\caption{The numbers under columns marked ``clean" are  classification accuracies (\%). The numbers under columns marked ``poison" are the backdoor attack success rates (\%)}
\caption{``Clean'' shows classification accuracy and ``Poison'' shows attack success rate.}
\label{table:results} 
\end{table*}

\subsection{MNIST}
The architecture of $\mathcal{G}_n$ is: $512\to160$, ReLU, $160\to160$, ReLU, and $160\to10$. The results are shown in Table~\ref{table:results} case a) and b). Our approach reduces the attack success rate to a low value with a small drop of classification accuracy. Neural Cleanse, however, mis-identifies $\mathcal{F}_b$ as a non-backdoored network for case a). For case b), Neural Cleanse detects multiple attacker chosen labels and therefore we give oracular knowledge to Neural Cleanse of the correct attacker chosen label(s). Fine-Pruning has high attack success rate on both cases. STRIP has high False Acceptance Rate (FAR) on both cases. 
%A well performed STRIP should have low FAR and low False Reject Rate (FRR). 
Therefore, for MNIST cases, our approach outperforms the prior works.

\subsection{GTSRB}
The architecture of $\mathcal{G}_n$ is: $512\to64$, ReLU, $64\to64$, ReLU, and $64\to43$. The results  are shown in Table~\ref{table:results} case c), d), and e). For case c) and d), our approach reduces the attack success rate to $1.5\%$ and $0$, respectively, whereas other approaches are not efficient. For e), our approach has a high attack success rate, but we will show, in our later experiment, that the attack success rate can be further reduced to $7.75\%$ with data augmentation, as shown in Table~\ref{table:aug}. In the GTSRB cases, our approach outperforms other approaches by having consistently low attack success rate.

\subsection{CIFAR-10}
The architecture of $\mathcal{G}_n$ is: $640\to160$, ReLU, $160\to160$, ReLU, and $160\to10$. The results are shown in Table~\ref{table:results} case f) and g). Neural Cleanse fails on both cases by either mis-identifying the $\mathcal{F}_b$ as non-backdoored network or having high attack success rate. Similarly, Fine-Pruning and STRIP fail on case g) by having high attack success rate or FAR. Our approach, however, reduces the attack success rate to almost zero, while maintaining high classification accuracy. Case f) was also used to test ABS \cite{LLTMAZ19}, wherein ABS mis-identified $\mathcal{F}_b$ as a clean network.
%The attack success rate of the reverse-engineered triggers made by ABS (RESAR) is $5.27\%$, which is far less than the $80\%$ threshold in the paper. Therefore ABS will mis-identify the backdoored network as non-backdoored network.

\subsection{YouTube Face}
The architecture of $\mathcal{G}_n$ is: $160\to1600$, ReLU, $1600\to4800$, ReLU, and $4800\to1283$. The results are shown in Table~\ref{table:results} case h), i), j), and k). Fine-Pruning fails on all cases by having high attack success rate. After using STRIP, the attack success rate is still high ($>10\%$). Neural Cleanse performs well only on case j) given oracular knowledge. For other cases, the attack success rate is high. Our approach, however, reduces the attack success rate in all cases, while maintaining reasonable classification accuracy. Note that the classification accuracy of our approach in this dataset drops more than in other datasets because its validation dataset is small and has only 9 images for each of the 1283 labels. 
%The validation dataset is not sufficiently representative of the training dataset.

\subsection{Comparison with RTLL}
We further compared our approach with the Re-Training the Last Layer (RTLL) \cite{ABCPK18} on some cases. The difference between RTLL and ours is that RTLL neither uses novelty detector nor changes the network structure. Our work shows that both changes are necessary. The results are shown in Table~\ref{table:tuning}.
%in \textbf{Supplementary Material}. 
Although RTLL decreases the attack success rate in case h) and i), the classification accuracy drops a lot. Additionally, RTLL fails on case d) and f). Our approach outperforms RTLL by reducing the attack success rate to almost 0 with a small drop of classification accuracy. 

\begin{table}
\begin{center}
\begin{tabular}{ccccc}
\hline\noalign{\smallskip}
& \multicolumn{2}{c}{RTLL} & \multicolumn{2}{c}{Our Approach} \\
Case  & Clean & Poison & Clean & Poison \\
\hline
d)  & 93.46 & 48.17 & 92.77 &  0.0 \\
f)  &  86.84 & 99.50 &  84.42 &  0.0 \\
h)  &  79.68 & 0.07 & 88.00 &  0.16 \\
i)  &  83.07 & $0.89$ & 88.35 &  $2.91$ \\
\hline
\end{tabular}
\end{center}
\caption{Comparison with RTLL.}
\label{table:tuning} 
\end{table}

\subsection{Robustness to Imperceptible Trigger}
We trained a backdoored network with small perturbations (only one pixel at each corner) as the trigger (i.e., the third $\mathcal{F}_b$ in CIFAR-10 case). The Badnet has $82.44\%$ classification accuracy and $91.99\%$ attack success rate. Our approach reaches $76.48\%$ classification accuracy and $4\%$ attack success rate, (i.e., our method still works even if the backdoor is imperceptible). 

\subsection{Large-Scale Dataset}
We also tested our method on the full ImageNet dataset (1000 classes) by creating a backdoored ImageNet model based on DenseNet-121. The dataset and trigger are shown in Fig.~\ref{fig:alltriggers} (e). $\mathcal{G}_n$ is simply a linear layer from 1024 to 1000. We observe that our defense reduces the attack success rate down to $12\%$ and the classification drops from $72\%$ to $63\%$. This is similar to the drop of classification accuracy on the YouTube Faces dataset, which also has more than a 1000 classes (and therefore a small number of clean validation dataset per each label). However, ImageNet is a much larger dataset and more complex than YouTube Face. Thus, we conclude that the size and complexity of the dataset is not a limiting factor for our method, although the number of classes does impact classification accuracy. 

\subsection{Smaller Validation Dataset}
%The number of used clean validation data for training the defense model will effect our online defense's performance due to the fact that our defense model is trained with the validation dataset.  
Table~\ref{table:R1Q2} shows the performance of our defense trained with different portions of the original validation datasets for some of our backdoored networks.  
%For case c), with $75\%$ and $50\%$ of the original validation set, the attack success rate increases to $10.27\%$ and $20.69\%$, while the classification accuracy remains roughly the same. For case d), both the classification accuracy and attack success rate remain roughly the same with $75\%$ and $50\%$ of the original clean validation dataset. For case f), with $75\%$ and $50\%$ of the original validation set, the classification accuracy remains roughly the same, whereas the attack success rate increases to $15.7\%$ and $37.6\%$, respectively. For case h) and i), the classification accuracy drops significantly with smaller validation datasets, while the attack success rate remains roughly the same. 
Smaller validation datasets are less representative of the training data distribution %(i.e., a domain shift happens)y that's actually not domain shift
and impact the performance of our approach.
While the quality and size of the validation dataset %Larger the extent of domain shift, the less 
impacts the effectiveness of our method,
%will be, but 
this is true for all existing defenses as well. %Addressing limited amounts of validation data more effectively will be one of our future work. 

\begin{table}
\begin{center}
\begin{tabular}{ccccc}
\hline\noalign{\smallskip}
 & \multicolumn{2}{c}{50 \% Validation Data } & \multicolumn{2}{c}{75 \% Validation Data } \\
%& \multicolumn{2}{c}{ } & \multicolumn{2}{c}{ }  \\
Case & Clean & Poison  & Clean  & Poison  \\
\hline
c) & 93.01 & 20.69 & 94.03 & 10.27  \\
d) & 91.83 & 0.01 & 92.62 & 0.03 \\
f) & 83.18 & 37.6 & 83.62 & 15.7  \\
h) & 76.35 & 0.35 & 83.18 & 0.11  \\
i) & 77.29 & $0.12$ & 83.52 & $0.17$ \\
\hline
\end{tabular}
\end{center} 
%\vspace*{-0.1cm}
\caption{Effectiveness of Our Approach on Smaller Dataset.}
\label{table:R1Q2}
\end{table}

\subsection{Training with Augmented Validation Data}
From Table~\ref{table:results}, although our approach significantly reduces the attack success rate for all the cases, the reduction is smaller for case (e) (only down to $69\%$). We observe that by adding standard Guassian noise to the clean validation dataset and training the novelty detector $\mathcal{N}$ and new decision function $\mathcal{G}_n$ with the augmented data, the attack success rate can be reasonably reduced, as shown in Table~\ref{table:aug}. The attack success rate is less than $6\%$ (only 8\% for case e) for all the cases with a small drop of classification accuracy. However, as we demonstrate next, the drop in accuracy can be mitigated if we update the novelty detectors online. 
%during online implementation with retraining, which will be shown in the following subsection.

\begin{table}
\begin{center}
\begin{tabular}{ccccc}
\hline\noalign{\smallskip}
 &  \multicolumn{2}{c}{No Data Augmentation} & \multicolumn{2}{c}{Data Augmentation} \\
Case    & Clean & Poison & Clean & Poison \\
\hline
a)  & 95.97 & 97.6  & 95.72 & 1.94\\
 b)   & 88.06 & 8.96  & 85.77 & 5.76\\
\hline 
 c)  & 94.96 & 1.50 & 92.60 & 2.56 \\
 d)  & 92.77 & 0.0 & 93.36 & 0 \\
 e) & 93.61 & 69.6  & 93.61 & \textbf{7.75}\\
\hline
f) & 84.42 & 0.0 & 80.04 & 1.35 \\
 g)   & 84.12 & 0.08 & 80.04 & 0.02 \\
\hline
 h)  & 88.00 & 0.16 & 84.54 & 0.24 \\
 i)  & 88.35 & 2.91 & 84.86 & 2.03 \\
 j)  & 87.78 & 1.9,0.03,5.6 & 84.07 & 0.08,0.03,4.5 \\
  k) & 87.74 & 0.5,0.01,0 & 85.13 & 0.07,0.01,0 \\
\hline
\end{tabular}
\end{center} 
%\vspace*{-0.1cm}
\caption{Data Augmentation.}
\label{table:aug}  
\end{table}

\subsection{Retraining with Poisoned Data}
%On-line implementation of our defense classifies the incoming stream of test inputs as either clean or poisoned.
%split the streaming input data into normal and poisoned groups. For the normal group, the output of the backdoored network is trusted. 
With the aid of a human expert to relabel the poisoned data, we can retrain a new decision function (i.e., only the classifier) on the poisoned test inputs as identified during the on-line implementation of our defense. 
We utilized the detected poisoned input in the first $20\%$ of the test data as well as the clean validation data for retraining. The remaining $80\%$ of the test dataset was used to evaluate the retrained models.  After retraining,  classification accuracy improves while maintaining low attack success rate (Table.~\ref{table:retraining}). We empirically observed that retraining with a small portion of the poisoned input data (10-20\%) was sufficient to achieve better classification accuracy removing the need to continually retrain the network. The online defense only works with  aid of a human expert labeler; nevertheless, our offline defense  outperforms the state-of-the-art. For the ImageNet case, if 8\% of the training dataset (50 images per label) is utilized as the validation dataset, the classification accuracy is $\approx 63\%$ and the attack success rate drops to $0.14\%$ with retraining.

\begin{table}
\begin{center}
\begin{tabular}{ccccc}
\hline\noalign{\smallskip}
 & \multicolumn{2}{c}{No Retraining} & \multicolumn{2}{c}{Retraining} \\
Case    & Clean & Poison & Clean & Poison \\
\hline
a)  & 95.98 & 1.98  & 96.60 & 0.39\\
 b)   & 87.78 & 2.68  & 88.1 & 1.68\\
\hline 
 c)  & 92.81 & 2.42 & 91.43 & 0.98 \\
 d)  & 93.30 & 0.0 & 93.94 & 0.02 \\
 e)  & 93.47 & 7.84  & 93.53 & 3.38 \\
\hline
f)  & 79.45 & 1.85 & 81.8 & 0.35 \\
 g)   & 79.97 & 0.02 & 83.62 & 0 \\
\hline
 h) & 84.31 & 0.21 & 88.65 & 0.07 \\
 i)  & 84.38 & 2.05 & 88.94 & 0.25 \\
 j)  & 84.27 & 0.08,0.03,4.5 & 89.25 & 0.07,0.02,0 \\
  k)  & 85.35 & 0.08, 0.02,0 & 88.83 & 0.08,0.06,0.1 \\
\hline
\end{tabular}
\end{center} 
%\vspace*{-0.1cm}
\caption{Retraining with Poisoned Data.}
\label{table:retraining} 
\end{table}

\section{Conclusions}

A novel feature-based anomaly detection with minimal assumptions on the backdoor attack was proposed. The approach requires only a small clean validation dataset and is computationally efficient. Several experiments were implemented and the performance was compared with state-of-the-art algorithms. The results show that our approach outperforms the state-of-the-art by achieving lower backdoor attack success rate on poisoned inputs while keeping high classification accuracy on clean inputs.  Additionally, our defense is formulated in terms of features and outputs rather than internal network structure and  therefore  applies to non neural network-based models. 

\newpage
\bibliography{main}

%\clearpage
\newpage
\section{Supplementary Material}
\label{sec:sup}
\begin{table}[ht]
\begin{center}
\begin{tabular}{cccccc}
\hline\noalign{\smallskip}
Layer &  Channels & Filter Size & Stride &  Act.\\
\noalign{\smallskip}
\hline
\noalign{\smallskip}
Conv2d & 1 $\to$ 16 & 5 & 1 &  ReLU \\
MaxPool & 16 & 2 & 2 &  / \\

Conv2d & 16 $\to$ 32 & 5 & 1 & ReLU \\
MaxPool & 32 & 2 & 2 &  / \\

Linear & 512 $\to$ 512 & / & /  & / \\
Linear & 512 $\to$ 10 & / & /  & / \\
\hline
\end{tabular}
\end{center}
\caption{$\mathcal{F}_b$'s Architecture for MNIST: case \textbf{a)}}
\label{table:mnist-aaa} 
\end{table}
\begin{table}[ht]
\begin{center}
\begin{tabular}{cccccc}
\hline\noalign{\smallskip}
Layer Type &Channels & Filter Size & Stride  & Act.\\
\noalign{\smallskip}
\hline
\noalign{\smallskip}
Conv2d & 1 $\to$ 16 & 5 & 1 &  ReLU \\
MaxPool & 16 & 2 & 2 &  / \\

Conv2d & 16 $\to$ 4 & 5 & 1 &  ReLU \\
MaxPool & 4 & 2 & 2 & / \\

Linear & 64 $\to$ 512 & / & / & ReLU \\
Linear & 512 $\to$ 10 & / & / &  / \\
\hline
\end{tabular}
\end{center}
\caption{$\mathcal{F}_b$'s Architecture for MNIST: case \textbf{b)}}
\label{table:mnist-cla} 
\end{table}
\begin{table}[ht]
\begin{center}
\begin{tabular}{cccccc}
\hline\noalign{\smallskip}
Layer & Channels & Filter & Str. & Pad. & Act.\\
\noalign{\smallskip}
\hline
\noalign{\smallskip}
Conv2d & 3 $\to$ 32 & 3 & 1 & 1 & ReLU \\
Conv2d & 32 $\to$ 32 & 3 & 1 & 0 & ReLU \\
MaxPool & 32 & 2 & 2 & / & / \\

Conv2d & 32 $\to$ 64 & 3 & 1 & 1 & ReLU \\
Conv2d & 64 $\to$ 64 & 3 & 1 & 0 & ReLU \\
MaxPool & 64 & 2 & 2 & / & / \\

Conv2d & 64 $\to$ 128 & 3 & 1 & 1 & ReLU \\
Conv2d & 128 $\to$ 128 & 3 & 1 & 0 & ReLU \\
MaxPool & 128 & 2 & 2 & / & / \\

Linear & 512 $\to$ 512 & / & / & / & ReLU \\
Linear & 512 $\to$ 43 & / & / & / & / \\
\hline
\end{tabular}
\end{center}
\caption{$\mathcal{F}_b$'s Architecture for GTSRB: case \textbf{c)}} 
\label{table:whitebox}  
\end{table}
\begin{table}
\begin{center}
\begin{tabular}{cccccc}
\hline\noalign{\smallskip}
Layer & Channels & Fil. & Str. & Pad. & Act.\\
\noalign{\smallskip}
\hline
\noalign{\smallskip}
Conv2d & 3 $\to$ 192 & 5 & 1 & 2 & ReLU \\
Conv2d & 192 $\to$ 160 & 1 & 1 & 0 & ReLU \\
Conv2d & 160 $\to$ 96 & 1 & 1 & 0 & ReLU \\
MaxPool & 96 & 3 & 2 & 1 & Drop(p) \\

Conv2d & 96 $\to$ 192 & 5 & 1 & 2 & ReLU \\
Conv2d & 192 $\to$ 192 & 1 & 1 & 0 & ReLU \\
Conv2d & 192 $\to$ 192 & 1 & 1 & 0 & ReLU \\
MaxPool & 192 & 3 & 2 & 1 & Drop(p) \\

Conv2d & 192 $\to$ 192 & 3 & 1 & 1 & ReLU \\
Conv2d & 192 $\to$ 192 & 1 & 1 & 0 & ReLU \\
Conv2d & 192 $\to$ 10 & 1 & 1 & 0 & ReLU \\
AvgPool & 10 & 8 & 1 & 0 & / \\ \hline
\end{tabular}
\end{center}
\caption{$\mathcal{F}_b$'s Architecture for CIFAR-10. $p = 0.5$.}
\label{table:squarecircle} 
\end{table}
\begin{table}
\begin{center}
\begin{tabular}{ccccc}
\hline\noalign{\smallskip}
Layer & Channels & Filter & Stride & Act.\\
\noalign{\smallskip}
\hline
\noalign{\smallskip}
Conv2d & 3 $\to$ 20 & 4 & 1  & ReLU \\
MaxPool & 20 & 2 & 2  & / \\
Conv2d & 20 $\to$ 40 & 3 & 1  & ReLU \\
MaxPool & 40 & 2 & 2  & / \\
Conv2d & 40 $\to$ 60 & 3 & 1  & ReLU \\
MaxPool ($x_1$) & 60 & 2 & 2  & / \\
Conv2d ($x_2$) & 60 $\to$ 80 & 2 & 1  & ReLU \\
Linear ($y_1$) & 1200 ($x_1$) $\to$ 160 & / & / &  / \\
Linear ($y_2$) & 960 ($x_2$) $\to$ 160 & / & /  & / \\
Add & $y_1+y_2$ & / & / &  ReLU \\
Linear & 160 $\to$ 1283 & / & / & /  \\
\hline
\end{tabular}
\end{center}
\caption{$\mathcal{F}_b$'s Architecture for YouTube}
\label{table:sunglasses}
\end{table}

\end{document}